\documentclass[a4paper, 10pt, conference]{ieeeconf}      % Use this line for a4
\usepackage{FG2024}

\usepackage{amsmath}
\usepackage{amssymb}
\usepackage{booktabs}
\usepackage{rotating}
\usepackage{multirow}
\usepackage{hhline}
\usepackage{pifont}
\usepackage{adjustbox}

\usepackage[utf8]{inputenc}
\usepackage{fourier} 
\usepackage{array}
\usepackage{makecell}

\FGfinalcopy % *** Uncomment this line for the final submission
\usepackage{pifont}
\newcommand{\xmark}{\ding{56}}%

\IEEEoverridecommandlockouts                      
                                                       
\def\FGPaperID{22} % *** Enter the FG2024 Paper ID here

\title{\LARGE \bf
Rethinking the Domain Gap in Near-infrared Face Recognition
}

\author{\parbox{16cm}{\centering
    {\large Michail Tarasiou$^1$ and Jiankang Deng$^1$  and Stefanos Zafeiriou$^1$}\\
    {\normalsize
    $^1$ Imperial College London}}
}

\begin{document}

\ifFGfinal
\thispagestyle{empty}
\pagestyle{empty}
\else
\author{Anonymous FG2024 submission\\ Paper ID \FGPaperID \\}
\pagestyle{plain}
\fi
\maketitle

%%%%%%%%%%%%%%%%%%%%%%%%%%%%%%%%%%%%%%%%%%%%%%%%%%%%%%%%%%%%%%%%%%%%%%%%%%%%%%%%
\begin{abstract}
Heterogeneous face recognition (HFR) involves the intricate task of matching face images across the visual domains of visible (VIS) and near-infrared (NIR). While much of the existing literature on HFR identifies the domain gap as a primary challenge and directs efforts towards bridging it at either the input or feature level, our work deviates from this trend. We observe that large neural networks, unlike their smaller counterparts, when pre-trained on large scale homogeneous VIS data, demonstrate exceptional zero-shot performance in HFR, suggesting that the domain gap might be less pronounced than previously believed. By approaching the HFR problem as one of low-data fine-tuning, we introduce a straightforward framework: comprehensive pre-training, succeeded by a regularized fine-tuning strategy, that matches or surpasses the current state-of-the-art on four publicly available benchmarks. Corresponding codes can be found at \verb|https://github.com/michaeltrs/RethinkNIRVIS|.

\end{abstract}

%%%%%%%%%%%%%%%%%%%%%%%%%%%%%%%%%%%%%%%%%%%%%%%%%%%%%%%%%%%%%%%%%%%%%%%%%%%%%%%%
\section{INTRODUCTION}
\label{sec:intro}

Face recognition (FR) is one of the most important and well-studied fields in computer vision \cite{zhao2003face,chellappa1995human}. It was for many years one of the main driving forces for the development of new lines of research in machine learning and was one of the first wins of Deep Neural Networks (DNNs) versus human perception \cite{schroff2015facenet}. Nowadays, FR technologies are widely adopted from cell-phones Face ID sensors to border control and immigration to name just a few. The most adopted and used systems currently operate with NIR images due to their high robustness to illumination changes. 

Heterogeneous face recognition (HFR) \cite{sun2013hybrid,light_cnn,adv_cross_spectral,dvg,dvg_face} is becoming essential in modern FR systems. While Near-Infrared (NIR) sensors are frequently used to capture face images during deployment, these images (probes) often need to be compared to a pre-existing face database (gallery) captured in the Visible (VIS) spectrum. Therefore, there's a pressing need for systems to effectively match faces across NIR and VIS modalities, highlighting the importance and growing interest in HFR. Most published HFR works suggest the presence of a domain gap as one of the main challenges in HFR \cite{dvg,dvg_face} and propose techniques to bridge that gap. 

\begin{figure}[t]
    \hspace*{-0.25cm} % This moves the figure 1 cm to the left
    \makebox[\linewidth][c]{%
        \includegraphics[width=0.85\linewidth]{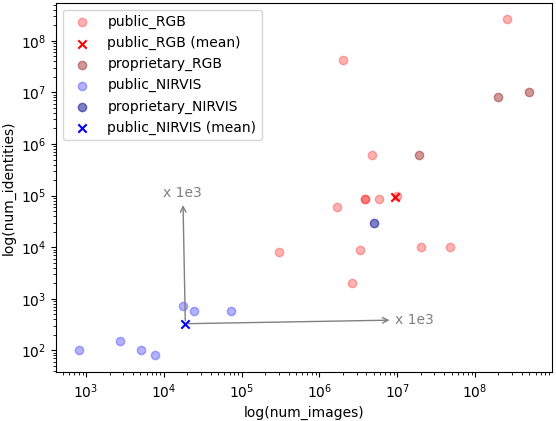}
    }
    \caption{Size of FR datasets ($\#$images, $\#$identities). The average size of NIR-VIS datasets is three orders of magnitude smaller than RGB datasets.}
    \label{fig:datasets}
\end{figure}

We follow a fundamentally different approach. Motivated by the perceptual similarities between VIS and NIR imagery (Fig.\ref{fig:ir_spectrum}) and the richness of VIS FR datasets (Fig.\ref{fig:datasets}) we employ transfer learning for solving the HFR problem. Our main observations and contributions are the following:

\begin{figure}[t]
   \centering
   \includegraphics[width=0.925\linewidth]{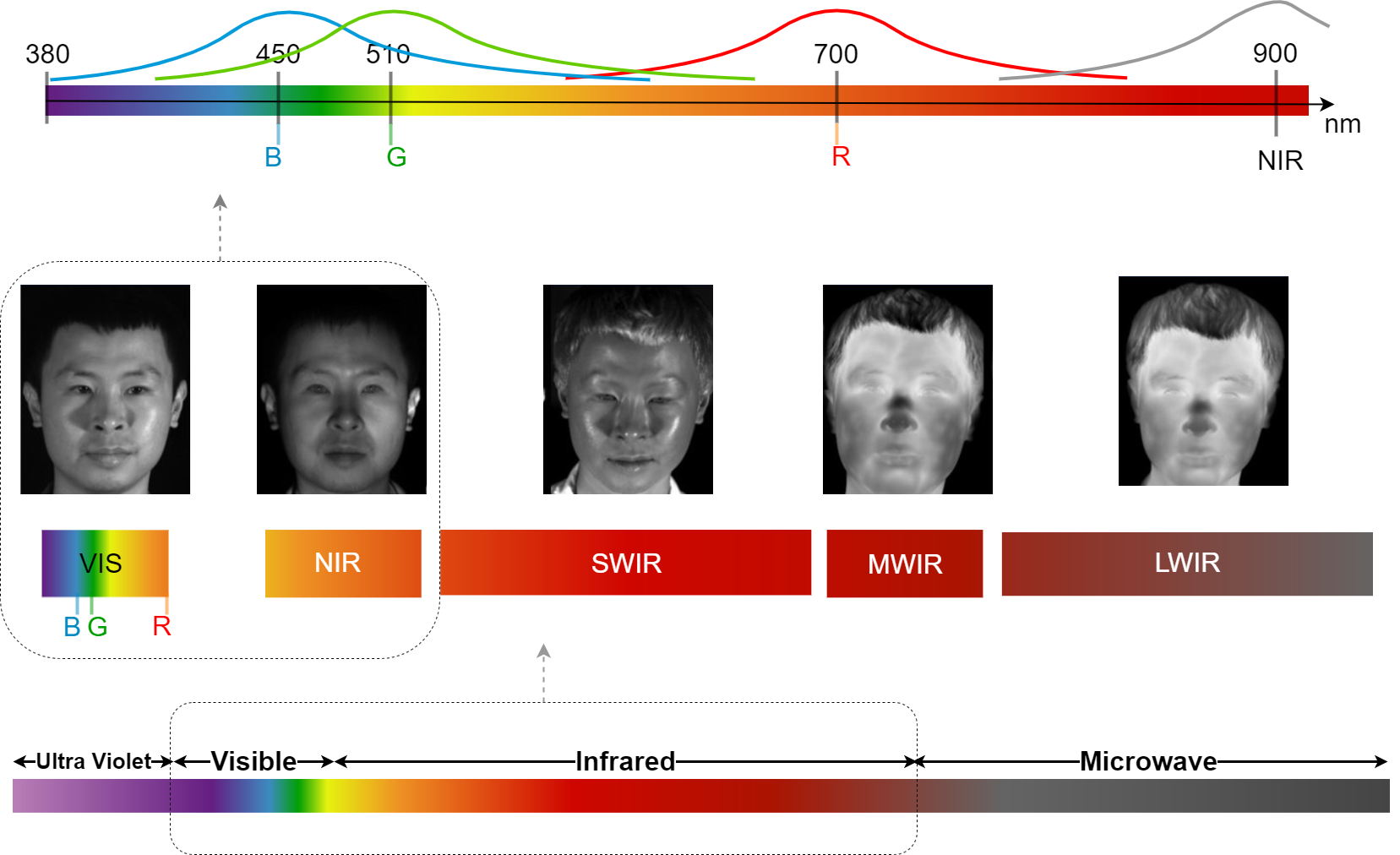}
   \caption{Face photo captured under visible and infrared light \cite{vis_nir_imagery_of_a_subject}. The infrared spectrum can be divided into four sub-bands: NIR (0.75–1.4$\mu m$), SWIR (1.4–3$\mu m$), MWIR (3–8 $\mu m$), and LWIR (8–15$\mu m$) \cite{ir_spectrum}. The spectral sensitivity of NIR imagery is much closer to that of the VIS spectrum opposed to images captured at the far end of the IR spectrum.}
    \label{fig:ir_spectrum}
\end{figure}

\begin{enumerate}
    \item \textbf{Domain gap}: we have determined that large CNNs, when pre-trained on extensive VIS data, show remarkable zero-shot performance in NIR-VIS HFR, even outperforming current benchmarks. This observation contrasts the prevailing HFR narrative of a large domain gap and has been missed by the HFR literature which has focused exclusively on training smaller models that do not exhibit this behaviour.
    \item \textbf {VIS pre-training}: based on the above finding, we shift our focus towards harnessing large-scale VIS data for HFR and introduce pre-training strategies which lead to demonstrably improved zero-shot performance.
    \item \textbf{NIR-VIS fine-tuning}: standard fine-tuning is found to disrupt the embedding space developed during pre-training. A simple method is presented that does not only rectify previous issues but also sets new performance benchmarks on four public NIR-VIS HFR datasets. Furthermore, through harnessing large-scale VIS data during fine-tuning, we find further improvements in sensor generalization performance.
\end{enumerate}

\section{RELATED WORK}
\label{sec:related}

\textbf{Primer on face recognition.} Recent years have witnessed a number of advancements in deep face recognition \cite{DeepFace,FaceNet,A-SoftMax,cosface,ArcFace}, the majority of which are based on the evolution of training loss functions. Most of the early works rely on metric-learning based loss ~\cite{chopra2005learning,FaceNet}, however, these methods are usually inefficient on large-scale training datasets, suffering from the combinatorial explosion in the number of sample combinations. Therefore, research attention has moved to margin-based classification loss functions that aim to enhance intra-class compactness and inter-class separability \cite{wen2016discriminative,A-SoftMax,AM-SoftMax,cosface,ArcFace}.
\textbf{NIR-VIS heterogeneous face recognition. }
There are two dominant approaches in the modern deep HFR literature:
1) {\bf Image synthesis} methods propose to solve the HFR problem by bridging the domain gap at the level of model inputs, by learning to translate faces across domains \cite{facesynthesis1,facesynthesis2,adv_cross_spectral}. 
2) {\bf Domain-invariant feature learning} methods \cite{domain_invariant4,domain_invariant5,domain_invariant1} aim at extracting facial identity features which are invariant to the source image domain, thus, bridging the domain gap at the level of extracted features. 
Among these, \cite{dvg, dvg_face} choose an unconditional generative model trained to generate paired NIR-VIS images from random noise and generate a large amount of training samples which are used to train a network to learn a domain invariant feature space. 
To the best of our knowledge, the current state-of-the-art in HFR is achieved by \cite{physically_based}, who reconstruct 3D face shape and reflectance from a large 2D facial dataset and transform the VIS reflectance to NIR reflectance in order to generate large-scale photorealistic data in the NIR and VIS spectra for further fine-tuning. 
\textbf{Transfer learning} aims at improving a learner's performance on a target task and data domain pair by ``transferring'' the knowledge already learned through training in different but somehow related source task and domain pair \cite{pan_transfer}. 
Transfer learning through reusing classifier weights has been extensively used as a means for knowledge distillation \cite{Chen_2022_CVPR} including works on FR \cite{Deng_2019_ICCV}. However, transfer learning for FR typically involves transferring to a different set of identities
which discards the possibility of reusing classifier weights.
To avoid this issue, \cite{local_adaptive} pre-compute the classifier as the mean per-class embedding of the pre-trained backbone and freeze these values to fine-tune the backbone for homogeneous FR. Additionally, they do not allow model parameters to deviate significantly from pre-trained values through an L2 regularization term.  

\begin{figure}[t]
    \centering
    \includegraphics[trim={0 0 0 2.5cm},width=0.9\linewidth]{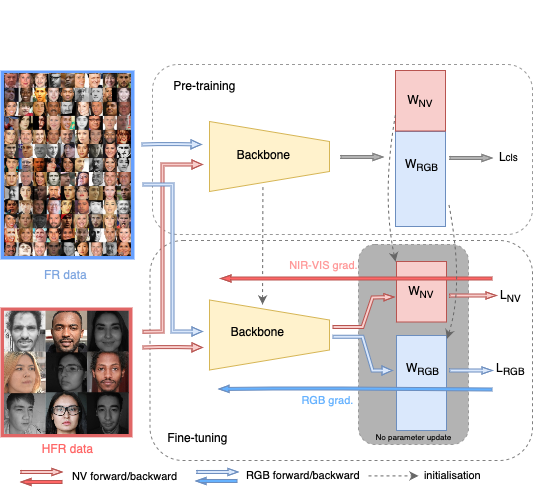}
    \caption{Proposed pre-training and fine-tuning with a subspace classifier for HFR. (top) we utilize both {\it source} and {\it target} data, use augmentation from Eq.(\ref{eq:augmentation}) and train with a joint set of identities, (bottom) we initialize all modules from pre-trained counterparts, feed both {\it source} and {\it target} data to our backbone, freeze both linear classifier weights, and train with the combined loss presented in Eq.(\ref{eq:finetune_loss}).}
    \label{fig:method}
\end{figure}

\section{METHOD} \label{sec:method}
In contrast to VIS images, the use of NIR cameras is not ubiquitous, discarding the possibility of gathering large-scale NIR imagery data from the public domain. This showcases the important role of large-scale VIS data as a source of pre-training data. 
A schematic overview of the proposed framework is presented in Fig.~\ref{fig:method}. 

\subsection{Pre-training with Large Scale VIS Data}
To achieve strong HFR performance a model needs to be able to achieve feature invariance for both VIS and NIR modalities. Current FR models trained on large-scale VIS datasets have arguably achieved very strong performances \cite{ArcFace,cosface}. Thus, we assume that pre-training on large VIS data is enough to learn a robust embedding space for the VIS modality and focus our attention on improving downstream transfer ability with regard to NIR images. Each face image can be decomposed into three color channels $x = \{x^{R}, x^{G}, x^{B}\}$ each of which is an intensity map of captured light at each respective spectral range. However, not all (R, G, B) channels share the same similarities with the NIR channel, the spectral sensitivity of the R channel has significantly higher overlap with the NIR spectral range than the B, G channels as shown in Fig.\ref{fig:ir_spectrum}. Motivated by this observation we are using the {\it red} channel as a means of shifting VIS images closer in appearance to the NIR spectrum through the following augmentation: 

\begin{equation}\label{eq:augmentation}
    x = \{ (x^{R}, x^{G}, x^{B}), (x^{R}, x^{R}, x^{R})\},  p=0.5
\end{equation}

Furthermore, we can optionally combine the {\it source} (VIS) and {\it target} (NIR-VIS) data for pre-training. In doing so we not only inject some {\it target} data knowledge during pre-training but also obtain a classifier checkpoint containing information about {\it target} identities which can be utilized directly during fine-tuning. 

\subsection{Fine-Tuning on Target NIR-VIS Data}
Fine-tuning DNNs directly for downstream tasks has been shown to potentially reduce performance in low data regimes \cite{kumar2022finetuning}, an observation which is also verified in section \ref{sec:experiments_finetuning}. While a pre-trained backbone transfers significant prior knowledge, FR classifier weights are typically initialized randomly and trained together with the backbone despite potentially having a larger capacity. We propose two techniques for transferring knowledge for FR classifiers.
First, given the strong zero-shot performance of VIS pre-trained models, it is reasonable to assume that the encoded representations of NIR-VIS data will also form compact clusters, the centers of which are expected to be strong identity predictors. We thus employ the mean identity embeddings \cite{local_adaptive} as classifier values for HFR. 
Second, assuming both  {\it source} and {\it target} data are available, we pre-train with both datasets and keep only the subspace of the classifier that corresponds to {\it target} identities. In doing so our {\it target} class centers fit well with respective identities and by explicitly comparing them with {\it source} centers during pre-training we end up with a more robust {\it target} embedding space. 
In both cases, we change the regularization scheme employed in \cite{local_adaptive}. Since there is a domain gap between {\it source} and {\it target} data we opt for a regularization scheme that does not penalize deviation from pre-trained parameter values. Instead, we reuse {\it source} data during fine-tuning and learn a simultaneously good solution for both HFR and homogeneous FR while placing no explicit constraint on model parameters.  

\begin{equation}\label{eq:finetune_loss}
    L_{finetune} = L_{cls}^{NIR-VIS} + \lambda L_{cls}^{preVIS}
\end{equation} 

\section{Experiments}

\begin{table}[t!]
\footnotesize
\begin{center}
\caption{FR and HFR datasets used in experiments.}
\label{tab:data}
\resizebox{\linewidth}{!}{
\begin{tabular}{cccccc}
Database   & Domain &$N_{images}$ & $N_{subjects}$ {\it (eval)} & $Year$ \\
\hline\hline
Oulu-CASIA \cite{oulu_casia} & NIR-VIS & 7,680  & 80 (40)  & 2009 \\
BUAA \cite{buaa} & NIR-VIS & 2.7k  & 150 (40) &  2012 \\
CASIA 2.0 \cite{casia_nirvis} & NIR-VIS & 17.5k & 725 (358) &  2013  \\
LAMP-HQ \cite{lamp-hq} & NIR-VIS & 73.6k & 573 (273) &  2019  \\
\hline
MS1Mv3 \cite{ms1m, ms1mv3}& VIS & 5.1M  & 93k   & 2020  \\
\end{tabular}
}
\end{center}
\end{table}

\begin{table}[t!]
\footnotesize
\begin{center}
\caption{Backbone Architectures used in experiments.}
\label{tab:models}
\resizebox{\linewidth}{!}{
\begin{tabular}{cccc}
Model   & input size & params (M) & FLOPS (G) \\
\hline\hline
MFN \cite{chen2018mobilefacenets} & $112\times112$ & 10.48 & 0.23 \\
LC29 \cite{light_cnn} & $128\times128$ & 10.48 & 3.70 \\
IR18 \cite{iresnet} & $112\times112$ & 24.03 & 2.62 \\
IR50 \cite{iresnet} & $112\times112$ & 43.59 & 6.32 \\
IR100 \cite{iresnet} & $112\times112$ & 65.15 & 12.12
\end{tabular}
}
\end{center}
\end{table}

{\bf Datasets. } 
Information on the datasets used is presented in Table \ref{tab:data}. We use the MS1Mv3 dataset \cite{ms1mv3} for RGB pre-training and the respective folds from four publicly available HFR datasets for fine-tuning and evaluation. 
{\bf Data pre-processing. } Following common practice for FR, we obtain normalized face crops by aligning all faces to a pre-defined template \cite{A-SoftMax,ArcFace,cosface}, using five facial landmarks extracted by RetinaFace \cite{retinaface}. 
{\bf Models. }All employed models are presented in Table \ref{tab:models}. Out of these, LC29 \cite{light_cnn} has been explicitly proposed for HFR.
{\bf Training.} We employ ArcFace \cite{ArcFace} as the margin based FR loss. We pre-train for 24 epochs, batch size 512, $\lambda=0.1$. We fine-tune for 20 epochs of target data using m=0.6, starting with learn rate $10^{-4}$ which we decay by 0.1 at epochs 10, 15, 20, batch size 64 keeping batch size for source data at 512. All training takes place on $\times 8$ Nvidia V100 GPUs.

\subsection{Zero-shot performance from VIS pre-training}
We begin by assessing the zero-shot performance of RGB pre-trained FR models in HFR without further fine-tuning, presented in Table \ref{tab:results zero-shot}. It is observed that larger architectures (IR50, IR100) behave qualitatively differently from smaller ones, having very strong performance despite the domain shift in stark opposition to very clear performance degradation for smaller models. This finding suggests that there exists adequate information in large-scale FR datasets to bridge the domain gap to NIR, however, this is not typically observed due to the small model capacity typically used in previous studies. Our proposed method for enhanced pre-training through augmentation offers clear performance gains for smaller architectures and less so for larger models. Finally, we observe that including target data in the pre-train set is enough to bridge a significant portion of the performance gap between the zero-shot and fine-tuned models.

\begin{table*}[!h]
    \centering
    \footnotesize
    \caption{Zero-shot NIR-VIS performance after pre-training (TAR@FAR=$10^{-4}$). $\dagger$ fold-1, * with target train data.}
    \begin{tabular}{c|ccc|ccc|ccc|ccc}
    Model& \multicolumn{3}{|c|}{Lamp-HQ $\dagger$ } & \multicolumn{3}{|c|}{CASIA 2.0 $\dagger$} & \multicolumn{3}{|c|}{Oulu-CASIA} & \multicolumn{3}{|c}{BUAA} \\
    & base & \textit{+ red aug.} & \textit{+target*} & base & \textit{+ red aug.} & \textit{+target*} & base & \textit{+ red aug.} & \textit{+target*} & base & \textit{+ red aug.} & \textit{+target*}\\
    \hline 
     MFN  & 87.91 & {\bf 88.68} &  96.90 & 95.05 & {\bf 95.75} &  98.26 & 84.60 & {\bf88.36} &  92.75& 96.70 & {\bf 96.73} &  98.44\\
     LC29 & 84.93 & {\bf 86.37} & 98.17 & 95.84 & {\bf 95.97} & 99.49  & 89.09 & {\bf 89.41} & 93.51  & 96.19 & {\bf 96.24} &  99.03 \\
     IR18 & 93.04 & {\bf 93.23} &  98.92 & 97.88 & {\bf 98.76}  &   99.51 & 92.72 & {\bf 94.80} &  95.74 & 98.05 & {\bf 98.45} &  99.37\\
     IR50 & 99.03 & {\bf 99.16} & 99.84 & 99.89 & {\bf 99.90} & 99.97  & {\bf 99.52} & 98.76 & 99.61  & {\bf 99.84} & 99.61 & 100.0 \\
     IR100 & 99.60 & {\bf 99.65} &  99.89 & 99.93 & {\bf 99.97} & 99.98 & 99.82 & {\bf 99.87} & 99.75  & {\bf 99.92} & 99.81 &  100.0  \\
    \hline
    \end{tabular}
    \label{tab:results zero-shot}
\end{table*}

\begin{table*}[!h]
    \centering
    \footnotesize
    \caption{Performance on NIR-VIS public datasets (TAR@FAR=$10^{-4}$) after (top) naive (pre-train/fine-tune), (bottom) regularized fine-tuning with either mean or subspace classifier and regularization scheme (\cite{local_adaptive}, $\lambda$=\{0,1\}). $\dagger$ fold 1.}
    \begin{tabular}{c|c|ccc|ccc|ccc|ccc}
    & Model & \multicolumn{3}{|c|}{Lamp-HQ$\dagger$} & \multicolumn{3}{|c|}{CASIA 2.0$\dagger$} & \multicolumn{3}{|c|}{Oulu-CASIA} & \multicolumn{3}{|c}{BUAA} \\
    &  & \xmark/\checkmark & \checkmark/\xmark & \checkmark/\checkmark & \xmark/\checkmark & \checkmark/\xmark & \checkmark/\checkmark & \xmark/\checkmark & \checkmark/\xmark & \checkmark/\checkmark & \xmark/\checkmark & \checkmark/\xmark & \checkmark/\checkmark\\
    \hline
     \multirow{4}*{\rotatebox{90}{\bf Naive}} & MFN & 0.14 & 87.91& {\bf 92.75}& 1.61& {\bf 95.75}& 81.54& 3.28& {\bf 84.60}& 60.46& 0.15& 96.70& {\bf 97.58}\\
    & IR18 & 3.98& 93.04& {\bf 94.77}& 0.73& {\bf 97.88}&  86.10&8.20 & {\bf 92.72} & 54.94 & 0.19 & {\bf 98.05} & 93.80\\
    & IR50 & 68.47 & {\bf 99.03} & 97.72 & 50.50 & {\bf 99.89} & 94.30 & 12.59 & {\bf 99.52} & 95.1 & 90.81 & {\bf 99.84} & 99.61\\
    & IR100 & 71.52 & {\bf 99.60} & 96.85 & 52.35 & {\bf 99.93} &  93.86 &  4.08 & {\bf 99.82} & 92.76 & 89.80 & {\bf 99.92} & 99.77\\
    \hline
    
     & & RCT \cite{local_adaptive} & $\lambda$=0 & $\lambda$=1 & RCT \cite{local_adaptive} & $\lambda$=0 & $\lambda$=1 & RCT \cite{local_adaptive} & $\lambda$=0 & $\lambda$=1 & RCT \cite{local_adaptive} & $\lambda$=0 & $\lambda$=1\\
    \hline
     \multirow{4}*{\rotatebox{90}{\bf Mean }} & MFN & 99.12& {\bf 99.50} & 99.34 &99.52 & {\bf 99.61} & 99.58 & 94.05& 93.65 & {\bf 96.61} & 98.64& 99.23 & {\bf 99.52}\\
    & IR18 & 99.67 & {\bf 99.77} & 99.70 & 99.83 &  99.89 & {\bf 99.90} & 96.59 & 95.82 & {\bf 96.96} & 99.61 & {\bf 100.0} & 99.84\\
    & IR50 & 99.91 & {\bf 99.93} & 99.91 & {\bf 99.98} & {\bf 99.98} & {\bf 99.98} & {\bf 99.88} & 99.85 & {\bf 99.88} &{\bf 100.0} & {\bf 100.0}& {\bf 100.0}\\
    & IR100 & {\bf 99.93} & {\bf 99.93} & {\bf 99.93} & {\bf 99.98}& {\bf 99.98} & {\bf 99.98} & {\bf 99.97} & {\bf 99.97} & {\bf 99.97} & {\bf 100.0} & {\bf 100.0} & {\bf 100.0}\\
    \hline
     \multirow{4}*{\rotatebox{90}{\bf Subspace}} & MFN & 99.34 & {\bf 99.69} & 99.54  & 99.60  & {\bf 99.68}  &  {\bf 99.68} & 95.12  & 95.73  &  {\bf 97.16} &  {\bf 99.52} &  99.34  & 99.41  \\
    & IR18 & 99.73 & {\bf 99.78} & 99.76 & 99.86 & 99.90  & {\bf 99.92} & 96.58 & 96.78 & {\bf 99.45}  & 99.70  & 99.92  & {\bf 99.95 } \\
    & IR50 & 99.91 & {\bf 99.95} &  99.93 & {\bf 99.98} &  {\bf 99.98} &  {\bf 99.98} & 99.88 & 99.90  &  {\bf 99.96} & {\bf 100.0} & {\bf 100.0}   &  {\bf 100.0} \\
    & IR100 & 99.93 & 99.93 & {\bf 99.94}  & {\bf 99.98} &  {\bf 99.98} & {\bf 99.98}  &  {\bf 99.97} &  {\bf 99.97} &  {\bf 99.97} & {\bf 100.0 } &  {\bf 100.0 } & {\bf 100.0 } \\
    \end{tabular}
    \label{tab:results naive finetune}
      \vspace{-0.2cm}

\end{table*}

\begin{table*}[!h]
    \centering
    \footnotesize
    \caption{Comparison with state-of-the-art. LC29 architecture with mean embedding classifier and $\lambda=0$. Folds 1-10.}
    \begin{adjustbox}{width=2\columnwidth,center}
    \begin{tabular}{c|ccc|ccc|cc|cc}
    \multirow{2}*{Method} & \multicolumn{3}{c}{CASIA 2.0 $\dagger$} & \multicolumn{3}{|c|}{Lamp-HQ $\dagger$} & \multicolumn{2}{|c}{Oulu-CASIA} & \multicolumn{2}{|c}{BUAA} \\
    \cline{2-11}
    & FAR=$10^{-4}$ &$ 10^{-3}$ & Rank-1 & FAR=$10^{-4}$ & FAR=$10^{-3}$ & Rank-1 & FAR=$10^{-3}$ & Rank-1 & FAR=$10^{-3}$ & Rank-1\\
    \hline
    LAMP-HQ \cite{lamp-hq} & - & 98.2 ± 0.2 & 99.2 ± 0.0 & - & 78.2  ± 3.0 & 97.3 ± 0.2 & 89.0 & 100.0 & 93.4 & 98.8\\
    DFAL \cite{dfal} & - & 98.7 ± 0.2 & 99.1 ± 0.2 & - & - & - & 93.8 & 100.0 & 99.2 & 100.0\\
    OMDRA \cite{omdra} & - & 99.4 ± 0.2 & 99.6 ± 0.1 & - & - & - & 92.2 & 100.0 & 99.7 & 100.0\\
    DVG-Face \cite{dvg_face} & 99.2 ± 0.1 & 99.9 ± 0.0 & 99.9 ± 0.1 & - & - & - & 97.3 & 100.0 & 99.1 & 99.9\\
    LC-29 \cite{physically_based} & {\bf 99.90 ± 0.06} & {\bf 100.0} ± 0.0 & 99.9 ± 0.1 & 98.6 ± 0.4 & 99.4 ± 0.3 & 99.1 ± 0.3 & 99.1 & 100.0 & 99.8 & 100.0\\
  \hline
  LC-29 (ours) & {\bf 99.9 $\pm$ 0.1} & 99.95 $\pm$ 0.02 & {\bf 100.0} & {\bf 99.35 $\pm$ 0.2} & {\bf 99.87 $\pm$ 0.05} & {\bf 100.0} & {\bf 99.62} & {\bf 100.0} & {\bf 99.90}  & {\bf 100}\\
    \end{tabular}\label{tab:comparison with sota}
    \end{adjustbox}
      \vspace{-0.2cm}

\end{table*}

\renewcommand{\tabcolsep}{1.25pt}
\renewcommand{\arraystretch}{1.1} 
\begin{table}[!h]
  \footnotesize
  \caption{Cross dataset evaluation (TAR@FAR=$1^{-4}$). Pre-trained MFN is fine-tuned with mean classifier ($\lambda$=0 /$\lambda$=1). $\dagger$ Fold 1.}
  \centering
  \begin{tabular}{cc|c|c|c|c}
    \multicolumn{2}{c}{} & \multicolumn{4}{c}{\bf Evaluation} \\ 
    \multirow{5}*{\rotatebox{90}{\bf Training}} & &  Lamp-HQ $\dagger$ &  CASIA 2.0 $\dagger$&  Oulu-Casia &  BUUA  \\ 
    \cline{2-6}
&     Lamp-HQ $\dagger$& {\bf  99.50} / 99.34  &  99.17 / {\bf  99.35} &  85.57 / {\bf  92.81} & 92.91 / {\bf  97.66} \\ 
    \cline{2-6}
&     CASIA 2.0 $\dagger$& 88.30 / {\bf  91.63} & {\bf  99.61} / 99.58 &  82.35 / {\bf  92.88}  &  91.27 / {\bf  98.28}  \\ 
    \cline{2-6}
&     Oulu-Casia & 74.79 / {\bf  87.37} &  84.49 / {\bf  96.88}  & 93.65 / {\bf  96.61} &  87.79 / {\bf  96.50} \\ 
    \cline{2-6}    
&     BUUA & 86.18 / {\bf  88.31} & 96.86 / {\bf  98.16} &  84.13 / {\bf  91.31}  &  99.23 / {\bf  99.52}  \\  
    \cline{2-6}
&     no fine-tune & 88.68 &  95.75 &  88.36 &  96.73 \\ 
    \cline{2-6}
  \end{tabular}\label{tab:cross dataset eval}
  
  \vspace{-0.4cm}
\end{table}

\subsection{HFR Fine-Tuning Performance}\label{sec:experiments_finetuning}
In Table \ref{tab:results naive finetune} (top) we present experimental results on \textbf{naively fine-tuning} to target HFR data through randomly initializing classifier weights and end-to-end training. 
More specifically, we evaluate model performance with or without pre-training or fine-tuning. We observe that without pre-training all models perform substantially worse than pre-trained counterparts, in particular, the smaller architectures fail to learn any discriminative features. Thus, pre-training appears to be crucial for learning useful representations from small HFR datasets. Additionally, naive target set fine-tuning appears to destroy the embedding space learned during pre-training and lead to performance degradation. This is always the case for IR50, IR100, and almost always for IR18 and MFN. In Table \ref{tab:results naive finetune} (bottom) we present experimental results for \textbf{regularized fine-tuning} methods. We observe clear performance gains as most models reach performances close to 100$\%$ for most datasets and are never found to degrade performance compared to no fine-tuning. We additionally find that regularization w.r.t. parameter values of pre-trained network (RCT) does not help and is almost always suboptimal compared to no regularization ($\lambda=0$). This can be explained by the NIR-VIS domain gap as RCT was proposed for homogeneous data. Our proposed regularization ($\lambda=1$) is found to be somewhat less performant for the more diverse datasets (Lamp-HQ and CASIA) but offers important gains for the less diverse ones (Oulu-Casia and BUUA). 
In most cases tested our subspace classifier outperforms mean embedding, albeit at the added cost of {\it target}-specific pre-training. 
Further benefits of our fine-tuning method can be observed in Table \ref{tab:cross dataset eval} where we perform \textbf{cross-dataset evaluation} among the four HFR datasets. Similarly, we note that apart from Lamp-HQ and CASIA, $\lambda=1$ outperforms $\lambda=0$ in every case, with very large performance differences in nondiagonal elements that have been trained and evaluated in different datasets. Lastly, in Table \ref{tab:comparison with sota} we present a {\bf comparison with state-of-the-art methods} for HFR. A LC29 model is pre-trained with red channel augmentation, no target data, and fine-tuned with a mean embedding classifier and $\lambda=0$. We observe similar performance for CASIA 2.0 and significant gains for all other datasets. Importantly, our framework is conceptually much simpler than competing methods which rely on expensive processes for generating synthetic data or employ complex architectures.

\section{Conclusion and Future Work}
\label{sec:conclusion}
In this paper, we presented a simple method consisting of strong pre-training, followed by regularized fine-tuning, that demonstrated robust performance in HFR. Our experiments further revealed that large-scale models, in particular, showcase significant zero-shot performances compared to their smaller counterparts. This suggests that VIS data alone carry ample information to effectively address the HFR problem. While knowledge distillation (KD) might seem like a natural progression given these findings, our initial experiments with this technique did not yield the anticipated results, which could be attributed to various factors, including the intricacies of the HFR problem. Future work might focus on refining KD techniques applicable to HFR.

\bibliographystyle{plain} % We choose the "plain" reference style
\bibliography{ms} % Entries are in the "refs.bib" file

\end{document}